\documentclass[runningheads]{llncs}
\usepackage{graphicx}
\usepackage{graphicx}
\usepackage{booktabs}
\usepackage{amsmath}
\usepackage{amssymb}
\usepackage{multirow}
\usepackage{bm}
\usepackage{bbding}

\def\Fig#1{{Fig.~\ref{fig:#1}}}
\def\Eq#1{{Eq.~\eqref{eq:#1}}}
\def\Table#1{{Table~\ref{tab:#1}}}

\usepackage{xcolor}
\usepackage[colorlinks, citecolor=blue]{hyperref}

\begin{document}
\title{LIDIA: Precise Liver Tumor Diagnosis on Multi-Phase Contrast-Enhanced CT via Iterative Fusion and Asymmetric Contrastive Learning}
\titlerunning{Precise Liver Tumor Diagnosis on Multi-Phase Contrast-Enhanced CT}

\author{Wei Huang\inst{1,2,3} \and
Wei Liu\inst{2,3} \and
Xiaoming Zhang\inst{2,3}\Envelope \and
Xiaoli Yin\inst{4} \and
Xu Han \inst{5}\and\\
Chunli Li\inst{2,3,4} \and
Yuan Gao\inst{2,3} \and
Yu Shi\inst{4} \and
Le Lu\inst{2} \and
Ling Zhang\inst{2} \and
Lei Zhang \inst{1}\Envelope\and
Ke Yan\inst{2,3}
}
% index{Wei Huang}
% index{Wei Liu}
% index{Xiaoming Zhang}
% index{Yin, Xiaoli}
% index{Xu Han}
% index{Chunli Li}
% index{Yuan Gao}
% index{Yu Shi}
% index{Le Lu}
% index{Ling Zhang}
% index{Lei Zhang }
% index{Ke Yan}

\authorrunning{W. Huang, W. Liu, et al.}

\institute{College of Computer Science, Sichuan University, 610065, Chengdu, China. \\
\and
DAMO Academy, Alibaba Group \\
\and
Hupan Lab, 310023, Hangzhou, China \\
\and
Department of Radiology, Shengjing Hospital of China Medical University,  110004, Shenyang, China\\
\and
Department of Hepatobiliary and Pancreatic Surgery, First Affiliated Hospital of Zhejiang University, 310006, Hangzhou, China\\
\email{zxiaoming360@gmail.com}; \email{leizhang@scu.edu.cn}; 
}
\maketitle              % typeset the header of the contribution
\renewcommand{\thefootnote}{\relax}
\footnotetext{\Envelope\hspace{0.5em}Corresponding authors. The work was done during W. Huang's internship at Alibaba DAMO Academy.}
\begin{abstract}
The early detection and precise diagnosis of liver tumors are tasks of critical clinical value, yet they pose significant challenges due to the high heterogeneity and variability of liver tumors. In this work, a precise LIver tumor DIAgnosis network on multi-phase contrast-enhanced CT, named LIDIA, is proposed for real-world scenario. To fully utilize all available phases in contrast-enhanced CT, LIDIA first employs the iterative fusion module to aggregate variable numbers of image phases, thereby capturing the features of lesions at different phases for better tumor diagnosis. To effectively mitigate the high heterogeneity problem of liver tumors, LIDIA incorporates asymmetric contrastive learning to enhance the discriminability between different classes. To evaluate our method, we constructed a large-scale dataset comprising 1,921 patients and 8,138 lesions. LIDIA has achieved an average AUC of 93.6\% across eight different types of lesions, demonstrating its effectiveness.
%We also test LIDIA on an external cohort of 828 patients and achieved a mean AUC of 89.3\%, showing a good generalization ability.
Besides, LIDIA also demonstrated strong generalizability with an average AUC of 89.3\% when tested on an external cohort of 828 patients.


\end{abstract}
\section{Introduction}
\label{sec:intro}
Liver is the largest solid organ in human body and plays a crucial role in various physiological functions. Meanwhile, it can be a common site for many malignant and benign tumors. According to global cancer statistics, liver cancer became the third leading cause of cancer death worldwide in 2020 \cite{bilic2023liver}. This high mortality rate is partially attributed to the late diagnosis of liver cancer, since patients with late-stage liver cancer discovered have limited treatment options and often a poor prognosis as well. Therefore, early detection and accurate diagnosis of liver tumors have become an urgent clinical task. Dynamic contrast-enhanced computed tomography (DCE-CT) is a widely utilized imaging technology for the diagnosis of liver tumors. To obtain DCE-CTs, multiple images are scanned at consecutive time points after intravenous injection of contrast agents. These multi-phase images provide valuable diagnostic information about the characteristics (e.g.~vascularity) of lesions via the pattern of contrast agent enhancement~\cite{Marrero2014acg}. However, these characteristics may be difficult to interpret due to the high degree of diversity and heterogeneity of liver tumors, especially for rare tumor types and atypical imaging signs. Additionally, manual analysis of CT images is time-consuming and often influenced by personal experience and biases of radiologists~\cite{Ying2024NC}.

Previous studies \cite{Cao2023NM,Yan2023Plan,yao2022effective,Ying2024NC} have demonstrated the capability of deep learning technology in identifying subtle textural details and shape variations of tumors that are imperceptible to human observation. %Inspired by them, this study focuses on automatic detection and diagnosis of liver tumors with deep learning, offering rapid and accurate assistance to clinicians and refining the diagnostic process. However, this task is highly challenging. First, the imaging signs of liver tumors are complex, with different lesion types potentially exhibiting similar features and the same lesion type showing different features, which increases the difficulty for models to differentiate between categories. Second, there is a significant class and size imbalance problem in liver tumors. For example, in our collected dataset, the most common lesion types (cyst and metastasis) contain 30 times more samples than the least common type (FNH); The largest tumor occupies more than $10^6$ pixels while the smallest one occupies less than 10. This hampers the classification of rare lesion types and the detection of small lesions. 
In recent years, considerable efforts have been made on the segmentation, detection, and classification of liver tumors. A large proportion of works focus on the segmentation~\cite{bilic2023liver,Li2018HDense,Tang2020E2Net,Xu2021PA,Zhang2021NC,Hu2023LabelFree} or detection~\cite{Cheng2022HCC} of tumors without differentiating their types. These works proposed improved convolutional neural network (CNN) backbones~\cite{Li2018HDense}, novel losses using lesion edge information~\cite{Tang2020E2Net,Xu2021PA}, weakly-supervised teacher-student network~\cite{Zhang2021NC}, or synthetic training data~\cite{Hu2023LabelFree}. Some studies investigated methods to classify tumor types with manually drawn region-of-interests (ROIs)~\cite{Xu2022knowledge,Yasaka2018differ}. Two-stage methods~\cite{Zhou2021Frontiers,Ying2024NC} first detect tumor ROIs with Faster R-CNN, and then classify each ROI with 3D CNN. Recently, a 3D instance segmentation framework is proposed in~\cite{Yan2023Plan} to jointly segment and classify liver tumors. Besides, multi-phase image fusion has been studied using early fusion~\cite{Yan2023Plan}, hetero-modal image fusion~\cite{Cheng2022HCC}, ConvLSTM~\cite{Yao2022ConvLSTM}, spatial and channel attention~\cite{Ying2024NC}, etc.

Despite progresses have been made, there are still two issues that need to be addressed in real-world scenario. First, the clinical guideline for liver tumor diagnosis~\cite{Marrero2014acg} recommends that a liver DCE-CT includes arterial, venous, and delayed phases. However, in practice, the delayed phase is not always scanned. This is partially because delayed phase prolongs scan duration, and many liver tumors are found in abdominal DCE-CTs not specially designed for liver, thus do not include delayed phases. Most existing algorithms neglected the delayed phase~\cite{Ying2024NC,Yan2023Plan,Zhou2021Frontiers,Xu2022knowledge}, resulting in the omission of valuable information for diagnosis. Second, most existing algorithms only considered common lesion types% such as hepatocellular carcinoma (HCC), intrahepatic cholangiocarcinoma (ICC), metastasis (meta), hemangioma (heman), focal nodular hyperplasia (FNH), and cyst
. Meanwhile, in practice there are numerous less common tumor types that hold significant clinical importance as well and need to be differentiated from the common ones. These rare lesions may present with imaging features similar to other types of liver tumors, or they might occur infrequently, resulting in a lack of ample data for effective training and recognition.

% introduce algorithm and novelty
% Brief intro to data, results, and compare with baselines
% Can be applied to other multi-modality lesion segmentation problems
% Summarize contributions
To address these problems, we propose a precise LIver tumor DIAgnosis network for multi-phase contrast-enhanced CTs, named LIDIA. LIDIA iteratively fuses all available CT phases to comprehensively capture and analyze the characteristics of lesions at different time points. Additionally, we introduce an asymmetric contrastive learning approach to address the heterogeneity of indeterminate categories and rare lesions in real-world scenarios. To verify the effectiveness of LIDIA, we collected a large-scale contrast-enhanced CT dataset with 1921 patients, 2/3 of them with the delayed phase. 8,138 lesions of 8 classes were comprehensively annotated. Besides 7 common tumor classes, there is an ``others'' class including more than 20 relatively rare tumor types. LIDIA can not only effectively fuse multi-phase CT under incomplete phase conditions, but also accurately differentiate rare lesion types from common types. It achieves a mean classification AUC of 93.6\%, outperforming the widely used baseline models. We also test LIDIA on an external cohort of 828 patients and achieve a mean AUC of 89.3\%, showing good generalization ability.

\section{Method}
\subsection{Preliminaries}
\textbf{Problem Definition.}
We define our liver tumor diagnosis task as follows. Let $P = \{\text{NC}, \text{A}, \text{V}, \text{D}\}$ denote images of the non-contrast, arterial, venous, and delayed phases, respectively. Consider a dataset $\mathcal{D}$ composed of $N$ patient cases, with each case containing multi-phase CT images $\bm{X} = \{\bm{x}^{\text{NC}}, \bm{x}^{\text{A}}, \bm{x}^{\text{V}}, [\bm{x}^{\text{D}}]\}$. Here, the delayed phase image $\bm{x}^{\text{D}}$ is optionally included, as denoted by the brackets, to reflect its potential unavailability. Each case is paired with $K$ instance-level lesion masks $\{\bm{S}_{j}\}_{j=1}^{K}$ and the corresponding lesion classifications $\{\bm{C}_{j}\}_{j=1}^{K}$. Note that $K$ is the number of tumors of each case and may be different among cases. The objective is to develop a model $f: \bm{X} \to (\{\bm{S}_{j}\}_{j=1}^{K}, \{\bm{C}_{j}\}_{j=1}^{K})$ that utilizes the multi-phase images $\bm{X}$ to accurately predict the lesion masks and their associated classes. 
This model must be designed to effectively utilize the all available phases to accurately identify the lesions and predict the correct class of each lesion, accommodating the scenario in which the delayed phase may not be available for all cases.

\noindent
\textbf{Mask Transformer.} Recently, mask transformers have been proposed for various segmentation tasks. Rather than performing per-pixel classification as in traditional semantic segmentation methods, they predict a set of binary masks and assigns a class label to each mask, enabling instance segmentation~\cite{Cheng2021MaskFormer,Cheng2022Mask2Former,Zhang2023MPformer}. An example is Mask2Former~\cite{Cheng2022Mask2Former}, which uses a pixel encoder-decoder to generate multi-scale features and employs learnable embeddings as object queries. These queries interact with image features as well as themselves through a transformer decoder to segment objects and also identify their classes% through bipartite matching~\cite{Carion2020DETR}
. Benefiting from the masked attention mechanism, this approach allows the transformer block to concentrate on specific local regions where tumors are situated, making Mask2Former particularly effective for the segmentation and diagnosis of lesions, which are often small in size.

\subsection{Liver tumor diagnosis network (LIDIA)}
\begin{figure}[ht!]
	\centering
	\includegraphics[width=1\textwidth]{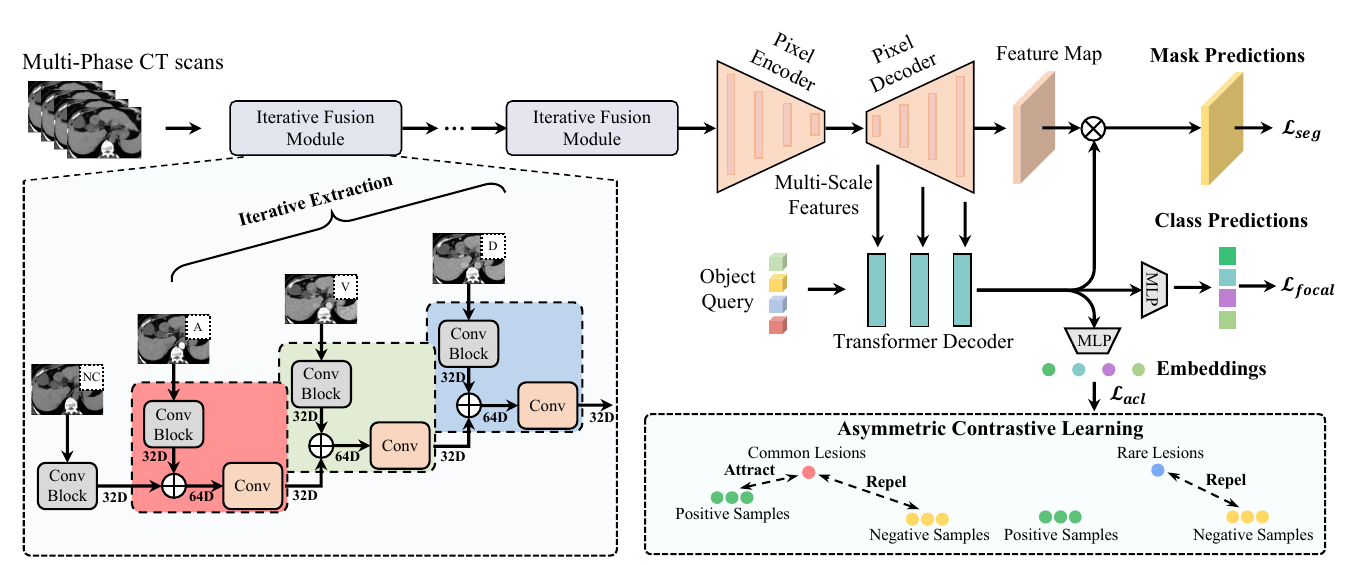}
	\caption{Illustration of the overall framework of LIDIA.}
	\label{fig:framework}
\end{figure}
We build LIDIA based on Mask2Former and introduce two key enhancements. First, we propose a multi-phase iterative fusion module, designed to handle multi-phase input and potential phase incompleteness. Second, we introduce an asymmetric contrastive learning loss to promote the discrimination between tumor types, especially for rare types. We also make a series of improvements in the training and inference procedure to enhance the robustness and accuracy of LIDIA. Our framework is shown in \Fig{framework}.

\noindent
\textbf{Iterative Fusion Module (IFM).}  In clinical workflows, physicians often determine the type of lesions based on the differences in features of lesion regions of interest (ROIs) across various phases. Inspired by this practice, to effectively utilize the complementary information between different phases, we first propose an iterative multi-phase fusion module. Formally, the specific information extraction for each phase can be defined as a function mapping from the phase image to feature map:
\begin{equation}
	\bm{h}^{p} = \mathcal{F}_p(\bm{x}^{p}) \in \mathbf{R}^C, \quad p \in P,
\end{equation}
where $\mathcal{F}_p$ represents the context-specific block for phase $p$, and $\bm{h}^{p}$ is the resultant feature map for that phase with $C$ channels. The feature extraction layer is comprised of two convolutional blocks, each employing 3D convolutions, instance normalization, and LeakyReLU activations to progressively extract information. Then, to effectively aggregate multi-phase information, we perform the fusion of information following the temporal sequence of the phases. Specifically, we fuse them in the order of non-contrast, arterial, venous, and delayed phases. %Starting with $\bm{h}^{init} = \bm{h}^{\text{NC}}$, 
The fusion process iteratively incorporates feature maps from each phase $p$ into a fused feature $\bm{h}^{fuse}$. The fusion operation is performed by concatenating the current fused  $\bm{h}^{fuse}$ with the feature of the next phase $\bm{h}^{i}$ and applying the convolutional block $\mathcal{F}_{conv}$ to extract new features. This is mathematically represented by the following recursive equation:
\begin{equation}
\bm{h}^{fuse}_{k+1} = \mathcal{F}_{conv}(\mathrm{concat}(\bm{h}^{fuse}_{k}, \bm{h})) \in \mathbf{R}^C, \quad \bm{h} \in \{\bm{h}^{p}\}_{p \in P}, \quad k \in \{1, 2, 3\},
\end{equation}
where $\bm{h}^{fuse}_1 = \bm{h}^{\text{NC}}$ and 
$\bm{h}^{fuse}_k$ represents the fused feature after incorporating the $k$-th subsequent phase's feature. %The order of phases used in the fusion process is such that $\bm{h}^{\text{NC}}$ precedes $\bm{h}^{\text{A}}$, which precedes $\bm{h}^{\text{V}}$, and finally $\bm{h}^{\text{D}}$. The final fused feature after processing all phases is denoted as $\bm{h}^{fuse} = \bm{h}^{fuse}_3$.
The final fused feature is $\bm{h}^{fuse}_4$ if delayed phase is present and $\bm{h}^{fuse}_3$ if not. This process provides a way to adaptively incorporate variable phase numbers. It is able to capture dynamic contrast changes among phases similar to ConvLSTM~\cite{Yao2022ConvLSTM}, while being more lightweight.
% With the described computational process, we have achieved not only the effective fusion of the non-enhanced, arterial, and venous phase images, but also provided a way to incorporate the potentially delayed phase images. This mimic the real-life clinical practice of capturing dynamic contrast changes.

\noindent
\textbf{Asymmetric Contrastive Learning (ACL).} To increase intra-class compactness and inter-class discriminability for liver tumors, we propose an asymmetric contrastive loss for the embedding of lesions, denoted as $\mathcal{L}_{acl}$. The liver contains numerous rare lesion types that are underrepresented (typically less than 10 samples in our dataset). It is impractical to assign each rare type as a separate class because the network could not effectively learn given such few samples, so we collectively assign as them as the ``others'' class. The ordinary supervised contrastive loss applies a uniform attraction within each class and repulsion between any two distinct classes, treating all classes equally. Due to the inherent heterogeneity of ``others'' class, clustering them using ordinary supervised contrastive loss would limit the model's flexibility when dealing with unseen lesions or those with similar appearances. On the other hand, we differentiates between common and rare lesions: $\mathcal{L}_{acl}$ only performs attraction within each common class and repulsion among common classes; while for samples within the ``others'' class, we do not try to cluster them together, but instead only keep them distant from the common classes. 
The original supervised contrastive loss can be represented by the following equation:
\begin{equation}
	\mathcal{L} = \mathbb{E}_{x\in\mathcal{I}} [\mathcal{L}(x)],
\end{equation}
where
\begin{equation}
	\mathcal{L}(x) =  \sum_{p\in\mathcal{P}(x)}
	-\frac{1}{|\mathcal{P}(x)|}
	\log\frac
	{\exp(z(x)^{T}z(p)/\tau)}
	{\sum\limits_{a\in\mathcal{A}(x)}\exp(z(x)^{T}z(a)/\tau)}.
	\label{eq:cl}
\end{equation}
Here, $z(x)$ is a non-linear projection function. $\mathcal{P}(x)$ is the set of positive samples with the same label as $x$. $\mathcal{A}(x)$ is the set of all contrastive samples with respect to $x$, and $\mathcal{I}$ is the set of training samples.

For common lesions, we perform intra-class attraction and inter-class repulsion using \Eq{cl}. For rare lesions, we simply push them away from the common classes; hence, there is no attraction amongst positive samples. Therefore, the contrastive loss becomes:
\begin{equation}
	\mathcal{L}(x_r) = \log
	{\sum\limits_{a\in\mathcal{A}_c(x_r)}\exp(z(x_r)^{T}z(a)/\tau)},
	\label{eq:rcl}
\end{equation}
where $\mathcal{A}_c(x_r)$ represents the set of common lesion samples with respect to $x_r$. Specifically, we project the object queries into an additionally embedding space to conduct above process. Meanwhile, to increase the sample size for contrastive learning, we perform cross-batch contrast via maintaining a memory bank.

\noindent
\textbf{Training and Inference.} Our task includes binary mask prediction and lesion type classification. For mask prediction, we employ a hybrid loss function, $\mathcal{L}_{seg}$, which combines Cross-Entropy (CE) loss with Dice loss. Similar to~\cite{Yan2023Plan}, we adopt a %method similar to mask2former, sampling $K$ pixels for the computation of segmentation loss, but we also implement a 
\emph{foreground-enhanced sampling strategy} when computing $\mathcal{L}_{seg}$, which increases the ratio of foreground pixels in the loss calculation, improving the recall of small lesions.%ensures that even the smallest lesions receive adequate representation and are accurately segmented. 
For lesion type classification, due to the class imbalance issue present among liver lesions, we employ \emph{focal loss} $\mathcal{L}_{focal}$~\cite{Lin2017FocalLoss} instead of the CE loss in~\cite{Cheng2022Mask2Former} to enhance the learning focus on the underrepresented classes. Therefore, the final loss function is expressed as a combination of above loss, mathematically represented as: 
\begin{equation}
\mathcal{L}_{final} = \lambda_1\mathcal{L}_{seg}+\lambda_2\mathcal{L}_{focal} + \lambda_3L_{acl}.
\end{equation}
Liver lesion classification faces the challenge of different tumors can present similar features while small lesions in segmentation tasks are easily missed due to their size. Therefore, when updating the model, we employ the \emph{sharpness aware minimization} strategy~\cite{Foret2020SAM}, which seeks weights that demonstrate low sensitivity to input noises. This leads to more robust predictions, especially for subtle or ambiguous features, thereby improving the model's performance on both classification and segmentation of lesions.

In the inference stage, our goal extends beyond achieving lesion-wise detection and pixel-wise segmentation. We also aim to obtain patient-wise diagnostic results, i.e.~the overall probability of the patient having each type of tumor. To achieve this goal, Zhu \textit{et al.}~\cite{Zhu2019S4C} suggested utilizing the size of the predicted lesion mask to infer the patient-level diagnosis. However, this method fails to consider the confidence of the mask prediction and tends to neglect small lesions. Yan \textit{et al.}~\cite{Yan2023Plan} adopted an additional network branch for patient-level classification, which is prone to overfitting compared to pixel-wise predictions. In this work, we propose a simple yet effective approach called \emph{LiverMax} to obtain patient-wise diagnosis probabilities from pixel-wise segmentation probabilities. We take the softmax output of the semantic segmentation result of LIDIA, compute the maximum value of each channel (tumor type) from all voxels inside the liver, and get the probability of the patient having each type of tumor. This strategy outperforms the previous strategies in our experiments.

\section{Experiments}

\textbf{Dataset.} We established a CT dataset comprising 1921 patients with 8,138 liver tumors annotated. Each patient underwent dynamic contrast-enhanced CT scans, with all patients having non-contrast (NC), arterial, and venous phases. 2/3 of them have delayed phase images. We registered all phases to the venous phase using DEEDS~\cite{Heinrich2013DEEDS}. Then, we invited a senior radiologist with 10 years of experience to delineate the tumors and annotate the types of all liver tumors based on pathological reports, imaging signs of CT and MRI, and follow-up information. Our study encompasses seven common types of liver lesions: hepatocellular carcinoma (HCC), intrahepatic cholangiocarcinoma (ICC), metastases (meta), hemangioma (heman), focal nodular hyperplasia (FNH), calcification (calc) and cyst. Additionally, we created a separate category for other rare lesions, which includes over 20 uncommon lesion types with each type typically fewer than 10 samples. We split the dataset into three subsets: 1305 samples for training, 298 for validation, and 318 for testing.

\noindent
\textbf{Implementation Details.} 
LIDIA is built upon of the nnU-Net~\cite{Isensee2021nnu} framework. The pixel-encoder is based on the encoder from the U-Net architecture, but the first layer have been replaced with IFM. For the decoder, LIDIA employs a Feature Pyramid Network. LIDIA is configured with 50 learnable queries. As for the loss weights, $\lambda_1=5$, $\lambda_2=5$, and $\lambda_3=0.01$. Additionally, the temperature is set as $0.1$, and the memory bank size is set as $1024$. During training, we use a batch size of $2$, a learning rate of 1e-5, and train the model for 1000 epochs.

\subsection{Experimental results}

\textbf{Comparisons with other methods.} We compare our proposed method with the widely-used robust baseline, nnU-Net~\cite{Isensee2021nnu}. Mask2Former~\cite{Cheng2022Mask2Former} achieved outstanding accuracy in instance segmentation of natural objects, thus we adapted it for 3D data and included it in the comparison. PLAN~\cite{Yan2023Plan} is a latest instance segmentation framework specially designed for liver tumor diagnosis. For nnU-Net and Mask2Former, patient-level results are inferred by counting the number of lesion pixels in their predicted masks as described in~\cite{Zhu2019S4C}. We report the patient-wise diagnosis (AUC-8: mean AUC of 8 tumor types; AUC-2: mean AUC of malignant and benign classification), lesion-wise detection (precision, sensitivity, and lesion classification accuracy), and pixel-wise lesion segmentation metrics in \Table{sota}. 
\begin{figure}[ht!]
	\centering
	\includegraphics[width=0.95\textwidth]{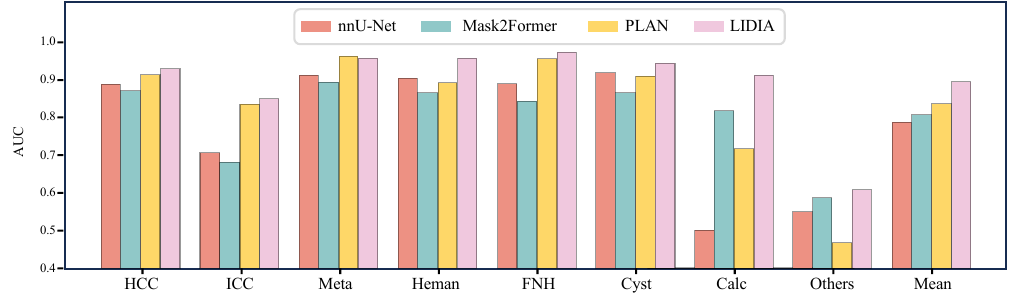}
	\caption{Illustration of AUC for all classes on an external cohort.}
	\label{fig:external}
\end{figure}
\begin{table}[ht!]
	\centering
	\caption{Comparisons with several state-of-the-art methods on the internal cohort.}
	\begin{tabular}{c|c|c|c|c|c|c}
		\hline
		\multirow{2}{*}{Method}                 & \multicolumn{2}{c|}{Patient-wise} & \multicolumn{3}{c|}{Lesion-wise} & Pixel-wise                                                         \\
		\cline{2-7}                             & AUC-8                            & AUC-2                           & Prec.          & Sens.          & Acc.            & Dice           \\
		\hline
		nn-UNet~\cite{Isensee2021nnu}           & 0.822                             & 0.915                            & \textbf{0.909} & 0.854          & 0.8575          & 0.873          \\
		Mask2Former~\cite{Cheng2022Mask2Former} & 0.841                             & 0.895                            & 0.832          & 0.861          & 0.8658          & \textbf{0.875} \\
		PLAN\cite{Yan2023Plan}                  & 0.905                             & 0.878                            & 0.907          & \textbf{0.886} & 0.8699          & 0.869          \\
		LIDIA                                   & \textbf{0.936}                    & \textbf{0.946}                   & 0.886          & 0.866          & \textbf{0.8812} & 0.869          \\
		\hline
	\end{tabular}%
	\label{tab:sota}%
\end{table}%
\begin{table}[ht!]
	\centering
	\caption{Comparison of the performance of various multi-phase fusion methods.}
	\begin{tabular}{c|c|c|c|c|c|c|c|c|c}
		\hline
		Method   & HCC            & ICC            & Meta           & Heman          & FNH            & cyst           & calc  & others         & Mean           \\
		\hline
		Baseline nnU-Net & 0.907          & 0.728          & 0.884          & \textbf{0.973} & 0.866          & 0.955          & 0.500 & 0.762          & 0.822          \\
		% Pad VP   & 0.908          & 0.732          & 0.888          & 0.963          & 0.866          & 0.954          & 0.500 & 0.799          & 0.826          \\
		HEMIS~\cite{Cheng2022HCC}    & 0.900          & 0.749          & \textbf{0.899} & 0.945          & 0.832          & 0.948          & 0.500 & 0.808          & 0.823          \\
		ConvLSTM~\cite{Yao2022ConvLSTM} & 0.890          & 0.715          & 0.890          & 0.972          & 0.900          & 0.935          & 0.500 & 0.781          & 0.823          \\
		IFM      & \textbf{0.913} & \textbf{0.757} & \textbf{0.899} & 0.961          & \textbf{0.902} & \textbf{0.956} & 0.500 & \textbf{0.815} & \textbf{0.838} \\
		\hline
	\end{tabular}%
	\label{tab:ifm}%
\end{table}%

\begin{table}[ht!]
	\centering
	\caption{Ablation study. IFM: iterative fusion module, ACL: asymmetric contrastive learning, SAM: sharpness-aware minimization}
	\begin{tabular}{c|c|c|c|c|c|c|c|c|c}
		\hline
		Method      & HCC            & ICC            & Meta           & Heman          & FNH            & cyst           & calc           & others         & Mean           \\
		\hline
		Baseline    & 0.861          & 0.711          & 0.857          & 0.935          & 0.797          & 0.880          & 0.790          & 0.753          & 0.823          \\
		+Focal Loss & 0.880          & 0.696          & 0.882          & 0.909          & 0.799          & 0.926          & 0.897          & 0.736          & 0.841          \\
		+LiverMax   & 0.934          & \textbf{0.853} & 0.947          & 0.974          & 0.968          & 0.942          & 0.912          & 0.826          & 0.919          \\
		+IFM        & 0.941          & 0.846          & 0.940          & 0.975          & 0.953          & 0.943          & 0.958          & 0.846          & 0.925          \\
		+ACL        & \textbf{0.954} & 0.845          & 0.948          & 0.973          & 0.966          & 0.946          & 0.937          & 0.856          & 0.928          \\
		+SAM(LIDIA) & 0.951          & 0.850          & \textbf{0.951} & \textbf{0.979} & \textbf{0.970} & \textbf{0.948} & \textbf{0.963} & \textbf{0.874} & \textbf{0.936} \\
		\hline
	\end{tabular}%
	\label{tab:ablation}%
\end{table}%

Firstly, LIDIA achieves the highest patient-wise AUC, which are the primary metric of focus in this work that are important for clinical diagnosis. For lesion-wise classification, LIDIA achieves the highest accuracy. % while balancing precision and sensitivity. In addition, despite LIDIA having only a lightweight decoder (FPN), it still achieves performance comparable to that of nnU-Net, which is specialized for segmentation task. 
The segmentation accuracy of all methods are comparable. To evaluate the generalizability of LIDIA, we test its performance on an independent external cohort. As shown in \Fig{external}, our method achieved optimal performance in almost all lesions, indicating good generalization capability. In summary, LIDIA achieves the highest diagnosis accuracy. We will display qualitative examples and lesion-wise confusion matrix in the Appendix.

\noindent
\textbf{Effect of iterative fusion module.} We used nn-UNet with three input phases (discarding the delayed phase) as the baseline method and compared the performance of various phase fusion approaches, where the AUC was calculated using the method described in~\cite{Zhu2019S4C}. The advantages of IFM are particularly evident across various lesions, including HCC, ICC, cysts, and ``others''. All methods except baseline nnU-Net can accept variable number of input phases and thus take advantage of the complete phase information (including delayed phase), yet IFM achieves better performance.

\noindent
\textbf{Ablation study.} As shown in Table~\ref{tab:ablation}, focal loss is beneficial for improving the underrepresented calc class samples. LiverMax effectively confines the foreground within the liver region, significantly reducing false positives and thereby increasing accuracy. Moreover, IFM efficiently exploits multi-phase information, substantially enhancing performance in heterogeneous categories such as HCC and others. Lastly, with the regularization and optimization methods, ACL and SAM, LIDIA achieves the highest average AUC of $93.6\%$. The findings from the ablation study clearly demonstrate the efficacy of our individual modules.

\section{Conclusion}
In this work, we introduce an effective approach to fuse multi-phase liver CT images to address the incomplete phase issue. Moreover, to minimize the impact of liver tumor heterogeneity on the model's classification performance, we propose an asymmetric contrastive loss. Our comprehensive evaluation on a large-scale dataset and external test set confirms the efficacy, generalizability, and clinical significance of our method.
\begin{credits}
\subsubsection{\ackname}
This work was partially supported by the National Natural Science Fund for Distinguished Young Scholars (No. 62025601), and partially supported by the National Natural Science Foundation of China (No. 82071885), the Innovation Talent Program in Science and Technology for Young and Middle-aged Scientists of Shenyang (RC210265), the General Program of the Liaoning Provincial Education Department (LJKMZ20221160), and the Liaoning Provincial Science and Technology Plan Joint Foundation."
\subsubsection{\discintname}
The authors have no competing interests to declare that are relevant to the content of this article. 
\end{credits}
\appendix
\begin{figure}[ht]
\centering
\includegraphics[width=1\textwidth]{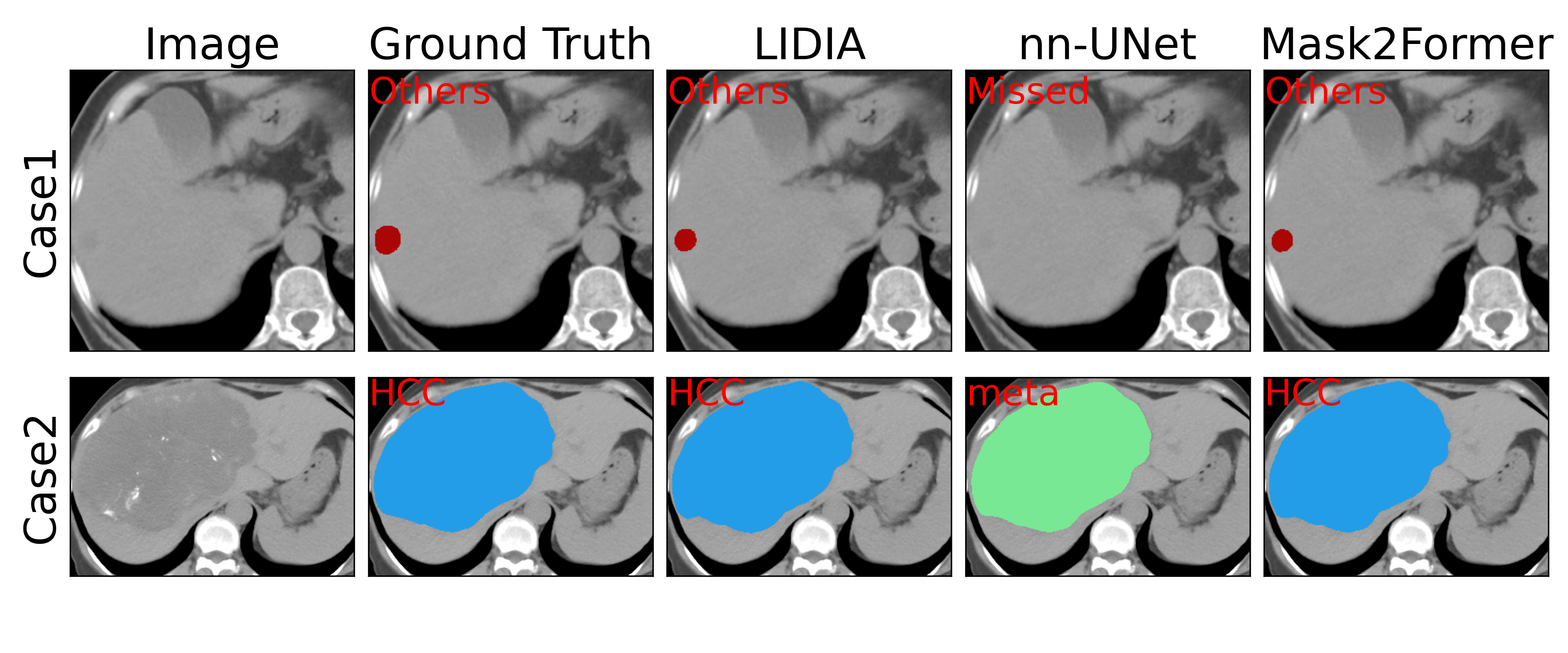}
\caption{Qualitative examples of lesion segmentation and classification in DCE-CT using
different methods.}
\end{figure}
\begin{figure}[ht]
\centering
\includegraphics[width=0.8\textwidth]{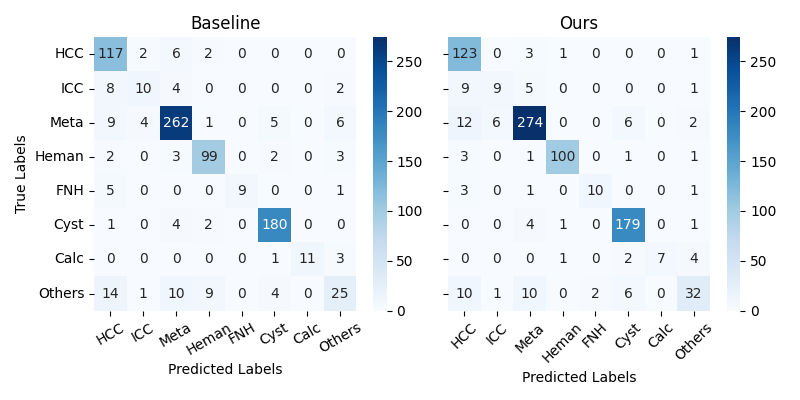}
\caption{Confusion matrix of lesion-level tumor classification in DCE-CT for recalled samples.}
\end{figure}
\newpage
\bibliographystyle{splncs04}
\bibliography{LatexSource-1629/Paper-1629}

\end{document}